\documentclass[sigconf]{acmart}

\AtBeginDocument{%
  \providecommand\BibTeX{{%
    \normalfont B\kern-0.5em{\scshape i\kern-0.25em b}\kern-0.8em\TeX}}}

\newcommand{\systemname}{\textit{NxtPost}}
\newcommand{\etal}{\textit{et al.}}

\usepackage{multirow}
\usepackage{adjustbox}
\usepackage{balance}
\usepackage{array}
\newcolumntype{R}[2]{%
    >{\adjustbox{angle=#1,lap=\width-(#2)}\bgroup}%
    l%
    <{\egroup}%
}

\copyrightyear{2021} 
\acmYear{2021} 
\setcopyright{acmlicensed}\acmConference[KDD '22]{Proceedings of the 28th ACM SIGKDD Conference on Knowledge Discovery and Data Mining}{August 14--18, 2022}{Washington DC, USA}
\acmBooktitle{Proceedings of the 28th ACM SIGKDD Conference on Knowledge Discovery and Data Mining (KDD '22), August 14--18, 2022, Washington DC, USA}
\acmPrice{15.00}
\acmDOI{10.1145/3394486.3403311}
\acmISBN{978-1-4503-7998-4/20/08}

\usepackage{natbib} 

\begin{document}

\title{NxtPost: User to Post Recommendations in Facebook Groups}
%% FEDOR: WE HAVE APPROVAL TO USE NXTPOST name (not nextpost)


\author{Kaushik Rangadurai, Yiqun Liu, Siddarth Malreddy, Xiaoyi Liu, \\ Piyush Maheshwari, Vishwanath Sangale, Fedor Borisyuk}
\affiliation{%
  \institution{Meta Platforms Inc.}
}

\renewcommand{\shortauthors}{Kaushik Rangadurai, \etal}

\begin{abstract}
In this paper, we present {\systemname}, a deployed user-to-post content-based sequential recommender system for Facebook Groups. Inspired by recent advances in NLP, we have adapted a Transformer-based model to the domain of sequential recommendation. We explore causal masked multi-head attention that optimizes both short and long-term user interests. From a user's past activities validated by defined safety process\footnote{Integrity violating posts filtered out from the data according to safety procedures \cite{GroupsIntegrity}.}, {\systemname} seeks to learn a representation for the user's dynamic content preference and to predict the next post user may be interested in. In contrast to previous Transformer-based methods, we do not assume that the recommendable posts have a fixed corpus. Accordingly, we use an external item/token embedding to extend a sequence-based approach to a large vocabulary. We achieve {\bfseries 49\%} abs.\ improvement in offline evaluation. As a result of {\systemname} deployment, 0.6\% more users are meeting new people, engaging with the community, sharing knowledge and getting support. The paper shares our experience in developing a personalized sequential recommender system, lessons deploying the model for cold start users, how to deal with freshness, and tuning strategies to reach higher efficiency in online A/B experiments.

\end{abstract}




\maketitle

\section{Introduction}\label{sec:intro}
\begin{figure}[tb]
  \vspace{0.5em}
  \centering
  \includegraphics[height=9cm,]{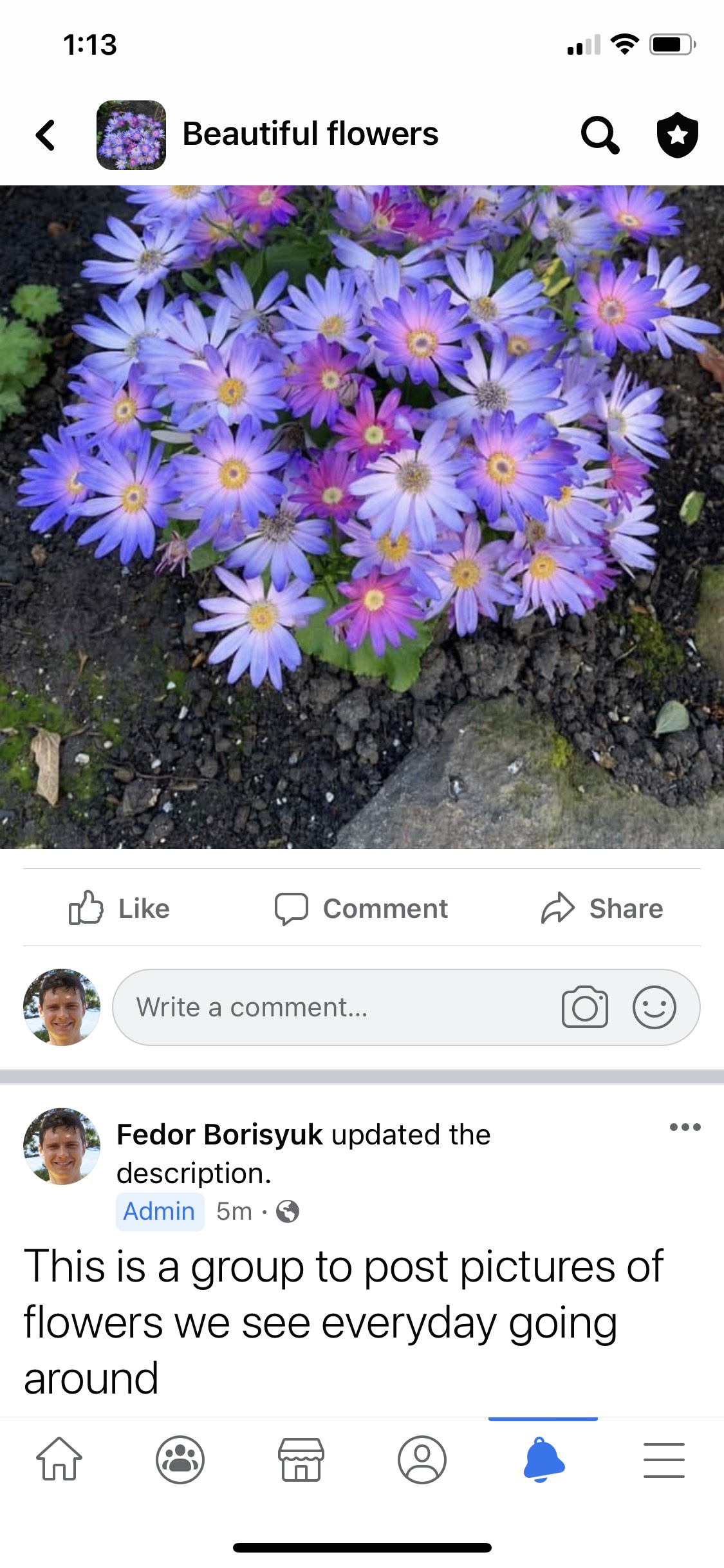}
  \caption{A screenshot of a Facebook Group about flowers.}
  \label{fig:marketplace_search_scr}
 % \vspace{-1em}
\end{figure}

Facebook Groups\footnote{https://www.facebook.com/groups} is a global platform that enables individuals with common interests to form communities and share their experiences~(Fig.~\ref{fig:marketplace_search_scr}). Hundreds of millions of people engage on the platform each day~\cite{fbgroup_stats}. Some groups may encompass conversations of diverse and general interests, whereas others cater to specific topics such as gaming, parenting, social learning, etc. Group members interact and share ideas by posting or commenting on a variety of content such as photos, videos, web links, and text.

To foster greater connection between individuals and communities, we are interested in developing a personalized recommender system for Facebook Groups. In particular, we aim to recommend publicly visible group posts to Facebook users for their enjoyment based on their respective content preferences. We formulate the objective as a sequential recommendation problem wherein a user's historical activity patterns are used in conjunction with static user features to predict the next group post that may likely interest the user. An activity history comprises dynamic content interactions such as likes, reactions, comments, and reshares. Examples of static user features include predicted language and home country. We call this recommender system {\systemname}.

There were several challenges in building {\systemname}:
\begin{itemize}
\vspace{-0.5em}
\itemsep0em
  \item \textit{Cardinality of posts:} Billions of posts are created each day with hundreds of millions of them engaged daily \cite{visrel_paper}. Therefore the cardinality of items to recommend pose a challenge in comparison to other production recommendation systems. While some try to solve the problem by recommending the top N most engaged posts, the cardinality is still in the order of hundred millions.
  
  \item \textit{Volatility of posts:} We observed that most posts have a short shelf-life, meaning that there is little overlap between engaged posts week over week. This is in contrast to most modern recommender systems which deal with relatively fixed recommendation set.

  \item \textit{Many types of engagement across multiple surfaces:} There are many forms of user engagement on Facebook that are explicit signals of relevance such as likes/reacts, shares, views, clicks, leaving a comment, or even liking a comment or commenting on a comment.  These user engagements also occur across different devices and surfaces/tabs within Facebook. Techniques that use engagement as a target signal have to contend with how to deal with different sources of relevance that are not directly comparable.
  %\vspace{-0.5em}
\end{itemize}

\begin{figure}[tb]
  \centering
  \includegraphics[width=\linewidth]{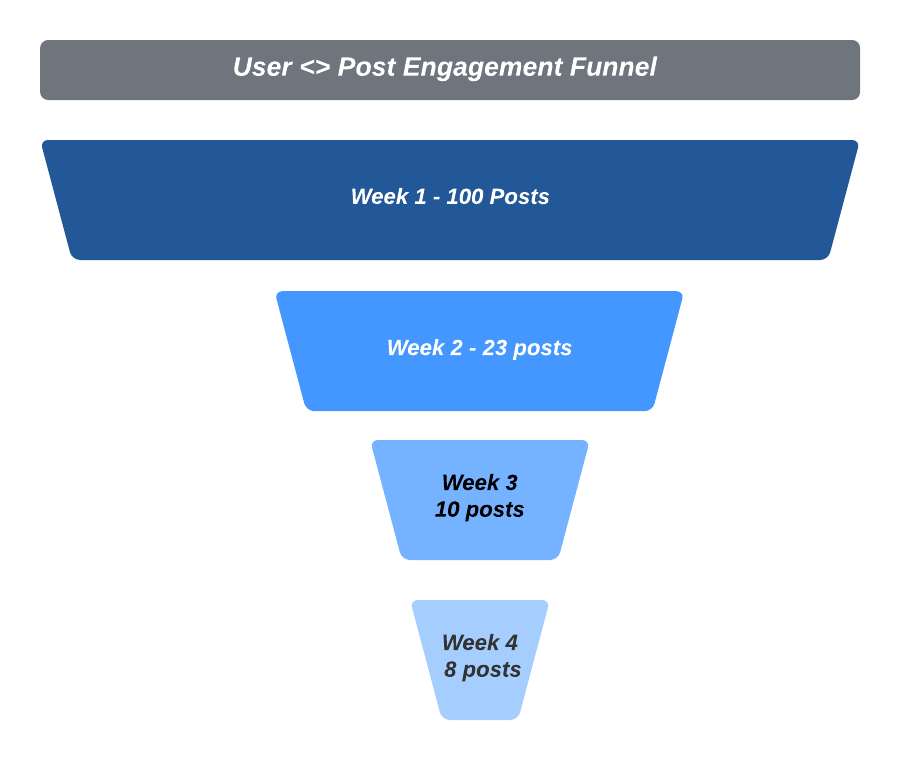}
  \vspace{-2em}
  \caption{\textbf{Posts have a short shelf-life}. Out of 100\% of posts engaged on the first week only 23\% continue to be engaged a week after and only 10\% of them 2 weeks after.}
  \label{fig:volatility_of_posts}
 % \vspace{-1.0em}
\end{figure}

To demonstrate the issue of volatility of posts we show how post engagement declines over time in Fig.~\ref{fig:volatility_of_posts}. Posts have a temporary nature; out of 100\% of posts engaged in a given week, only 23\% continue to be engaged the following week and only 10\% continue to be engaged two weeks later. We will next describe our model {\systemname} and how it addresses each of these aforementioned challenges.

This paper presents {\systemname}, a recommendation system that recommends posts to a user by predicting which post would likely come next in the sequence of the posts recently engaged by the user. We filter out integrity violating posts from the data according to defined safety procedures \cite{GroupsIntegrity}. Within {\systemname} we extend the idea of modeling sequences of words to the more general notion of modeling sequences of complex objects---in the form of posts---which contain text and multi-media. {\systemname} has demonstrated gains in production applications by introducing several modeling techniques described section \S\,\ref{sec:model}. In order to deal with a large vocabulary, we've removed the learnable token embeddings in the Transformer layer and replaced it with a pre-trained item embedding. We've also removed the classification layer and used a Two Tower architecture instead. We've also explored causal attention and 2 losses to optimize both the short and long-term user interests. We achieve over 49\% absolute offline metrics improvement in comparison to previous state-of-the-art modeling approaches. We share our modeling techniques in \S\,\ref{sec:model}, ablation studies in \S\,\ref{sec:experiments}, and production deployment experience and tuning tricks to achieve higher performance in online A/B experiments in \S\,\ref{sec:deployment}.

{\systemname} has been deployed to Facebook Groups; it powers user-to-post embedding-based retrieval. {\systemname} operates at Facebook Groups scale with hundreds of millions of users \cite{fbgroup_stats} consuming results of {\systemname} recommender system. As a result of {\systemname} deployment, 0.6\% more people are meeting new people, engaging with the community, sharing knowledge and getting support.

\section{Related Work}\label{sec:related_work}
\paragraph{Personalized Content Recommender Systems}

Collaborative filtering (CF) is a well-researched technique that has been adopted by a number of large-scale consumer applications~\cite{10.1145/371920.372071, 10.1145/1273496.1273596, 10.1109/MIC.2003.1167344, MLSYS2019_e2c420d9, kislyuk2015human, 10.1145/2959100.2959190}. CF-based recommender systems traditionally apply transductive learning on a $\left\langle \text{user}, \text{content} \right\rangle$ bipartite graph to identify content of interest to each user. However, if large number of new nodes are continuously added to the graph, the effectiveness of a CF-based system to suggest fresh and relevant content is markedly reduced by challenges related to cold-start~\cite{10.1145/2645710.2645751}. Thus, we opt for a content-based approach to empower a user-to-content recommendation system for Facebook Groups.

Graph Convolutional Network (GCN) has attracted industry attention in recent years~\cite{DBLP:conf/iclr/KipfW17}. \citet{10.1145/3219819.3219890} used graph structure to aggregate content representations for related-content recommendations and demonstrated improvement in both offline and online settings. A GCN-based system typically derives node representation from a weighted bag of neighboring embeddings. By contrast, in {\systemname} we take into consideration the position of each post in the user's interaction history and learn the weight of each post through attention. Further analysis on the correlation between future engagement actions and post positions is provided in \S\,\ref{sec:data}.

Transformers have been explored in the domain of Sequential Recommendation. \citeauthor{Kang2018SelfAttentiveSR} \cite{Kang2018SelfAttentiveSR} used causal attention that mimics a language model to learn the user representation while \citeauthor{10.1145/3357384.3357895} \cite{10.1145/3357384.3357895} used masked language model to learn a user representation. However, both these techniques assumed that the number of items in the Transformer layer were limited and doesn't change quickly. By contrast, {\systemname} uses combines a TwoTower approach with causal language modeling techniques to deal with a large dataset size at Facebook.

\paragraph{Sequential Modeling}

Two-tower neural networks have been widely used to model the semantic relevance between heterogeneous data types~\cite{huang2013learning, 10.1145/2988450.2988454, 10.1145/3366424.3386195, alibaba_two_tower}. \citet{10.1145/3298689.3346996} used recently watched videos as an input feature to the user tower, but the viewer history was represented by the average video ID embeddings instead of the watch sequence. \citet{10.1145/3308558.3313650} explored sequence encoding and showed improvement to recommendation quality by incorporating encoders of different temporal ranges. For {\systemname}, we provide an ablation study on sequence length in \S\,\ref{sec:sequence_length_and_layers}.

There have been some recent works on sequence modeling inspired by BERT. \citet{10.1145/3357384.3357895} proposed a BERT-like pre-training approach by randomly masking some items in the input sequences and then predicting the IDs of those masked items based on their surrounding context. In this paper, we share our experience in developing Transformer-based user-to-content models. In contrast to existing works, we extend the transformer usage for user-to-content prediction to unlimited vocabulary.  \citet{pinterest_workshop} proposed to use transformers with retrieval losses \cite{DBLP:journals/corr/JeanCMB14}, which improves performance of recommendation system further. We have observed improvements in model performance by adopting retrieval losses in our implementation (\S\ref{sec:transformer_encoder}).
In lieu of content IDs, we construct input features from pre-trained content embeddings and apply transformers to prepare user representations. We provide an ablation study of different model configurations in comparison with prior works (\S\,\ref{sec:experiments}). We report significant improvement in key metrics with {\systemname} over existing approaches.

\section{Modeling}\label{sec:model}

In this section, we describe the model architecture of {\systemname} and also explain how we collect training data and optimize the model. Central part of {\systemname} is a transformer encoder architecture with causal multi-head attention and support to input pre-trained item embeddings. We go over the details of these techniques and explain how we translate it to the domain of sequential recommendation.

\subsection{{\systemname} Model Architecture}

\begin{figure*} [!h]
\centering
\vspace{-0.1em}
\includegraphics[width=0.8\textwidth]{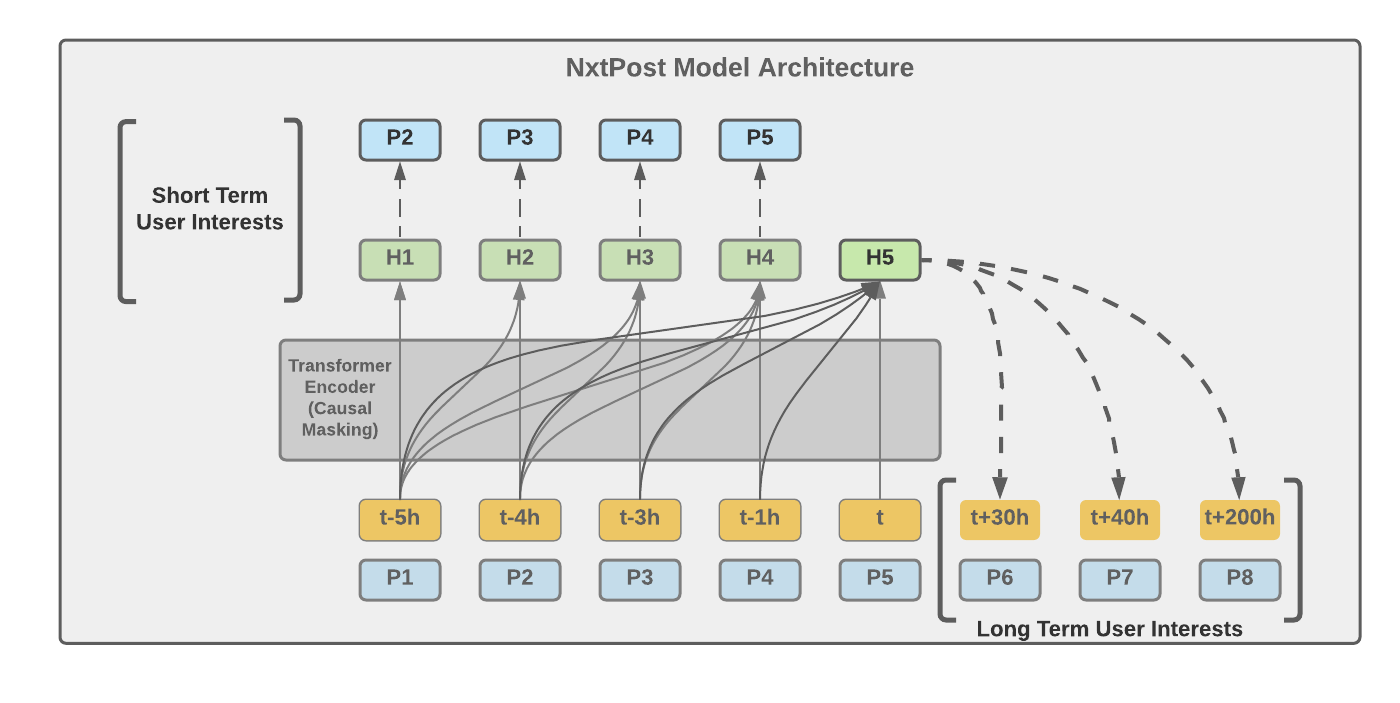}
\caption{Architecture of {\systemname}, optimizing for both users short-term and long-term interests.}
\label{fig:modelArchitectureFig}
% \vspace{-1.0em}
\end{figure*}

{\systemname} is a Two Tower / Dual Encoder architecture with in-batch negatives. It has a learn-able user tower and a fixed (pre-trained) post tower and has in-batch negatives. Before we get into the details of the model architecture, let us first understand what is a Two Tower architecture and how it works. In order to explain how two tower architectures work, we’ll explain what happens during training in a single batch of batch size B. In the TwoTower model, we apply the user tower obtain user embeddings tensor of size [B, D], and similarly we also obtain the post embeddings tensor of the same size. We then L2-normalize the embeddings and perform a matrix multiplication of the normalized embedding matrices. This will give us the logits matrix of shape [B, B]. This matrix represents user post similarity for every possible pair within a batch, with every row belonging to a user and every column belonging to a post. We treat this as a multi-class classification problem, where the number of classes is B, and ground-truth class indices lie in the main diagonal. We use multi-class cross-entropy loss to optimize our network. The main advantages of a TwoTower model over a classification model is that the number of learnable parameters in a TwoTower model is independent of the number of items/tokens in the vocabulary whereas a classification model depends on the number of number of items and in fact the final linear projection layer from the embedding dimension to the number of items in the vocabulary is often the bottleneck.

The model architecture is depicted in Figure \ref{fig:modelArchitectureFig}. Central part of the model is a Transformer encoder with causal masking. An important change that we make to the Transformer layer is to remove a learnable token embedding layer and replace it with pre-trained token embeddings. The reason for this is that, we've a large vocabulary (order of billions) and having an embedding layer of this magnitude is not feasible. We've also removed the classification layer and replaced it with a TwoTower architecture for the same reason. We've 2 losses - optimizing for users short-term and long-term interests. With the help of the causal masking, we make the hidden representation of the Transformer match at step t match with the post embedding at step (t+1). This helps learn the users short-term interests. In order to model the user's long-term interests, we take the final hidden representation and match it with multiple posts in the future. We use the final hidden representation from the Transformer layer as user's representation. We go over the various components of the model architecture in the sections below.

\subsubsection{Post/Item Embeddings}

For the post embeddings, we have features for a post that include text, multimedia such as images and videos, and additional metadata such as the poster's country and detected language. We use one shared 6-layer XLM-R \cite{DBLP:conf/acl/ConneauKGCWGGOZ20} encoder for all the textual fields. For each post there is a variable number of images attached to it. We use pre-trained image embeddings (\citet{visrel_paper}) for each image in the post we apply a shared MLP layer and use deep sets (\citet{DBLP:journals/corr/ZaheerKRPSS17}) fusion to combined the set of image embeddings into a fixed size representation. We then fuse the representations from different feature channels with learned attention weights to get the document's final embedding representation. For video  representation we used pre-trained video embeddings based on \citet{video_rep_paper3}. We train this model as a post to post similarity task using a Two Tower architecture. The architecture of the post tower is based on \citet{que2search_paper} and hence we'll not revisit this in detail here. The post embeddings are then fed as token embeddings to the Transformer encoder of the user tower, which we describe in the next section.

\subsubsection{User Tower}
\begin{figure}[tb]
  \centering
  %\vspace{-0.1em}
  \includegraphics[width=\linewidth]{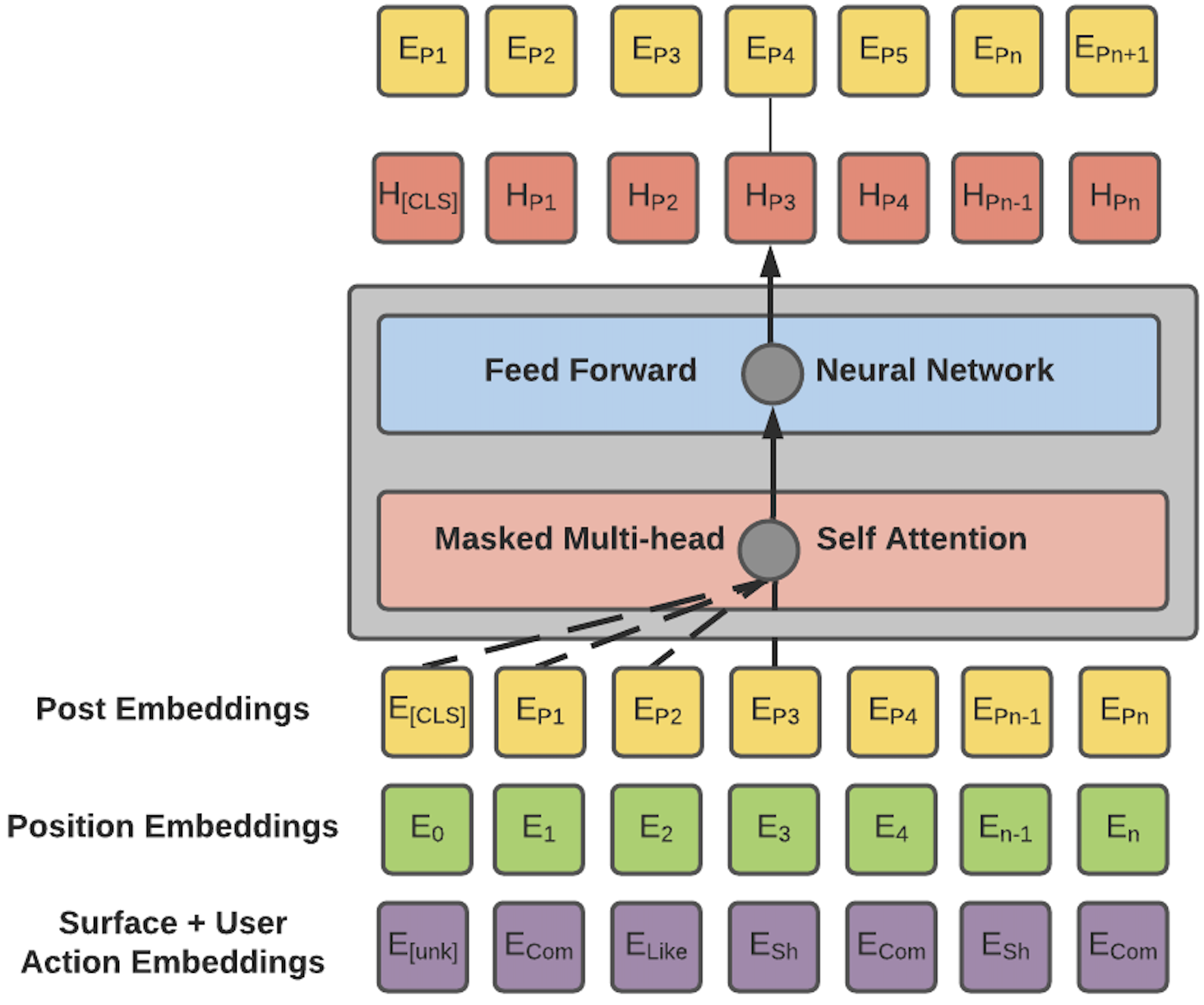}
  % \vspace{-2em}
  \caption{Causal pre-training - Given a prior set of posts, the model is required to predict the next post, which in this case is $P_4$. In a given batch of size $B$, we make upto $B * (max sequence length - 1)$ predictions in a batch.}
  \label{fig:gpt_pretrain}
 %  \vspace{-1.0em}
\end{figure}

For the user tower, we feed in a sequence of posts through a Transformer encoder layer with causal attention mask. While transformers usually take in a sequence of token ids and learn the token embeddings as part of the transformer layer, we feed the pre-trained post embeddings from the previous section. This is to deal with the large vocabulary of posts and the volatile nature of the relevance of posts. Along with the post embeddings, we also concatenate others feature like the time since current and treat the concatenated embedding as the embedding of the post. Additionally, we sum the post embeddings along with learned position embeddings and learned user action (like/comment/reshare) embeddings (see Fig.~\ref{fig:gpt_pretrain}) and feed this as input to the transformer layer.

\subsubsection{Short Term User Interests}

In order to model the user's short term interests, we take advantage of causal masking in Transformer encoder layer. For every time step t, we take the hidden representation of the Transformer encoder layer at a time step t and match it with the post embedding at time step (t+1). We use all the posts that belong to other users in the history as negatives and optimize using a Cross-entropy loss. The reason we need a causal encoder for this is that it avoids peeking into the future. By doing this, we're able to provide all prefixes of the history as training data to the model. This also makes the model more robust as the model is now optimizing for numerous (history, label) pairs in a single batch rather than just 1 slice of it.

\subsubsection{Long Term User Interests}

In order to model the user's long term interests, we take the final hidden representation from the transformer encoder with causal masking. We then match it with every post embedding at various time steps from (t+1) to (t+m) where m is the maximum number of labels we're considering. By doing this we're achieving two things - (i) we're able to model multiple user interests and (ii) we're also able to capture user's long-term interests. We use all the labels from other users as negatives and use Cross-Entropy loss to optimize the problem. The total loss is equal to the weighted combination of the short-term interest loss and the long-term interest loss.

\subsection{Transformer Encoder Layer}\label{sec:transformer_encoder}

We use a transformer encoder layer to encode the context sequence which has 4 building blocks: the embedding layer, multi-head attention, position-wise feed-forward network and a pooling layer at the end.

\subsubsection{Embedding Layer}

\begin{figure}[tb]
  \centering
  \includegraphics[width=\linewidth]{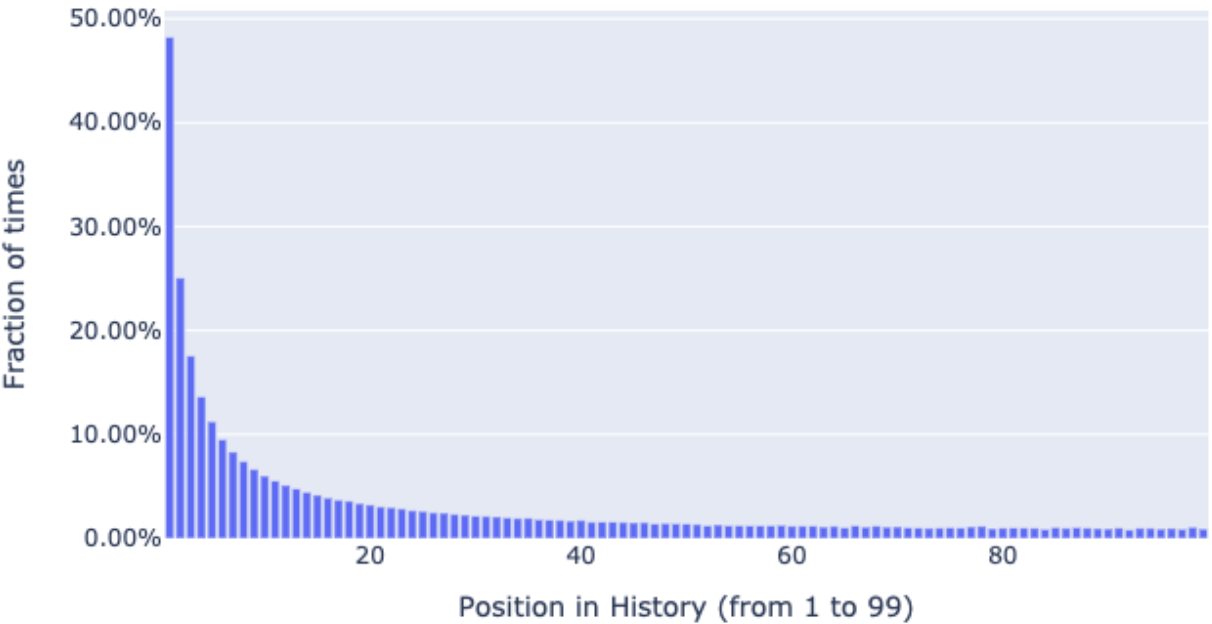}
 % \vspace{-2em}
  \caption{The sequence length data analysis. User's next engaged post in the future is mostly similar to several recently engaged posts from history.}
  \label{fig:seq_len_analysis}
  %\vspace{-1.5em}
\end{figure}

Traditionally a transformer layer has three embeddings: token embeddings, position embeddings and segment embeddings which are learned as part of the training process. From Fig.~\ref{fig:seq_len_analysis} we observe that last recent posts capture the most similarity with the post that would be engaged by the user in the future and hence we decided to keep the position embeddings. However, we needed to make changes to the token and segment embeddings. The token embeddings learn an embedding for every token in the language vocabulary, usually on the order of $10^5$s. Since posts have a short shelf-life and new posts are constantly being created it is impractical to learn an embedding for each post as part of the model. Instead of learning an embedding for each post we instead plug-in an external content-based post encoder that is trained on content similarity---two posts will have similar embeddings if their contents are similar. While segment embeddings make sense for natural language sentences (where each sentence belongs to a separate segment), we have replaced it with a combination of surface and user action embeddings. Intuitively, different user actions have a different weight (for example, commenting on a post might have a higher weight or importance than just viewing a post) and we learn an embedding for each of them. All the 3 embeddings are summed and passed to a transformer encoder layer. We've also added the [CLS] placeholder denotes the start of the sequence and its embedding is learned as part of the model.

\subsubsection{Multi-Head Attention.}

At the heart of the transformer context encoder is the multi-head attention. Attention was first popularized in sequence modeling and has since been widely adopted as it can capture the dependencies between items without regard to how close/far away they are from each other. In Multi-Head attention, we project the queries, keys and values $h$ times with different learned linear projections. On each of these projected versions of queries, keys and values, we perform attention in parallel yielding $d_v$ dimensional output values. These are then concatenated and once again projected resulting in the final values. 

\begin{equation}\label{pffn}
     \mathrm{Attention}(Q, K, V) = \mathrm{softmax}(QK^{T}/\sqrt{d_{k}})V
\end{equation}

\begin{equation}\label{head}
    \mathrm{head}_{i} = \mathrm{Attention}(Q{W_{i}}^{Q},K{W_{i}}^{K}, V{W_{i}}^{V})
\end{equation}

\begin{equation}\label{mha}
    \mathrm{MultiHead}(Q, K, V) = \mathrm{concat}(head_{1}, head_{2}, ... , head_{h})W^{o}
\end{equation}

\subsubsection{Position Wise Feed Forward Layer} Besides the multi-head attention, we apply a feed-forward network to each position separately and identically. This consists of 2 linear layers with a ReLU activation between them.

\begin{equation}\label{pffn}
    \mathrm{FFN}(x) = \mathrm{max}(0, xW_1 + b_1)W_2 + b_2
\end{equation}

\subsubsection{Pooling layer.} The final layer is a pooling layer, which takes a hidden representation at every position, pools them and provides a fixed representation of the entire sequence. We have explored mean pool, sum pool and attention pooling but found mean pool to work best.

% \begin{equation}
\begin{equation}\label{simple_attention_fusion}
\begin{aligned}
\varphi &= \{\varphi_i\}_{i=1}^N & \text{representations of N channels} \\
w &= \text{Softmax}(\text{proj}_N(\varphi_1\mathbin\Vert ... \mathbin\Vert \varphi_N)) & \text{$w$ is N channel weights} \\
f &= \sum_{i=1}^N w_i \dot \varphi_i & \text{final tower representation}
\end{aligned}
\end{equation}

We used PyTorch and several downstream libraries such as Facebook's PyText, Fairseq, and the Multimodal Framework (MMF) to implement the model. For all of our experiments, we used a dropout of 0.2, gradient clipping of 1.0, 2 transformer encoder layers, sequence length of 100, learning rate of $7 \times 10^{-4}$ with a batch size of 1024 and Adam optimizer.

\subsection{Training}\label{sec:data}
\begin{figure}[tb]
  \centering
  \includegraphics[width=\linewidth]{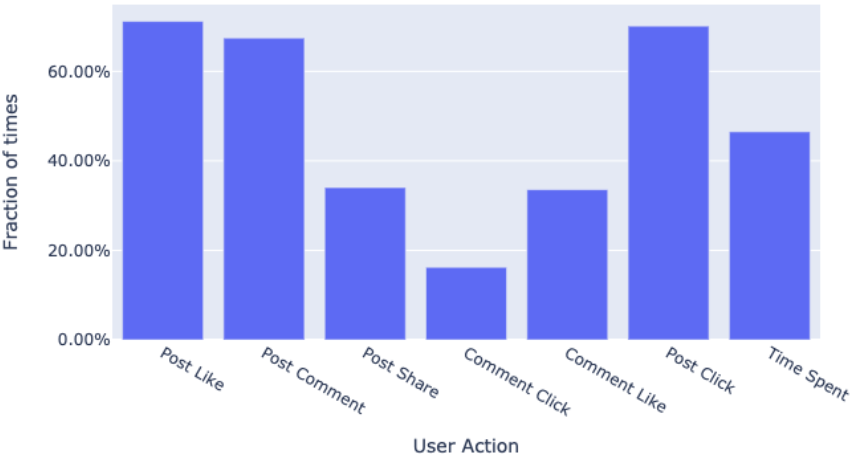}
  %\vspace{-2em}
  \caption{Are some actions more predictive than others? In short, yes.  Accuracy of predicting the next event in the sequence is much lower for comment click/like and post share than for other actions such as post like/comment/click and time spent (to a lesser extent).}
  \label{fig:relevant_actions}
  %\vspace{-1.0em}
\end{figure}

\textit{Training data.} We collected user actions on the posts across pages and groups in the Facebook ecosystem as positive examples, where the data is first de-identified and aggregated before training. We filter out integrity violating posts from the training data according to defined safety procedures \cite{GroupsIntegrity}. In order to find the correlation of user action to the probability that the user would engage with this post in the future, we analyzed many user contexts. More specifically, we split this context randomly and calculated the cosine similarity of the post embedding in the context with the target post. We then sliced it by the user action of the context post to Figure ~\ref{fig:relevant_actions}. We find that like, comment and post click are more correlated with the similarity of future posts than actions like comment click, comment like and comment react.

In our training setup, the dataset is positive (user, list of posts) pairs where in-batch negatives are used for negative examples. We have sampled batch negatives using techniques described in \citet{que2search_paper}. 
We've cleaned the data by taking the following measures: (1) filter-out posts with less than N interactions, (2) always consider only the history of user sequence and never look into the future.
We use a scaled multi-class cross-entropy loss (Eq.~\ref{scaled_cross_entropy}) to optimize our network, where $p_i$ denotes a post, and $u_i$ denotes a user, $cos$ denotes cosine similarity and $s$ denotes scale. The idea of having a scale times cosine is also mentioned in \citet{deng2019arcface}. During training, we found that having a $scale$ or temperature parameter is important for loss to converge. In our use case, we choose scale between 15 and 20
    \begin{equation}\label{scaled_cross_entropy}
     \textrm{loss}_{i} = -\log\frac{\exp(s \cdot \cos\{u_i, p_i\})}{\sum_{j=1}^{B} \exp(s \cdot \cos\{u_i, p_j\})} 
    \end{equation}

\subsection{Evaluation}\label{sec:eval_data}
Before running A/B tests online on live traffic, we first evaluate our model candidates offline and select the best one based on metrics such as batch Recall@K (also called Hits@k).

\textit{Batch Hits@K}: This metric measures whether diagonal element $\cos(q_i, d_i)$ is among the top K scores of the row $\cos(q_i, d_j)$, $j \in [1, B]$. 
This metric is easy to compute during training and is the closest metric that the model optimizes, enabling us to iterate fast on modeling ideas.
For evaluation, we hold out one day of data in the future respective to when the training data was collected and use it in the target post prediction task. In prediction we only consider the history of user sequence prior to the event of engaging with the target post and never look into the future.

\textit{KNN Hits@K or Hits@K}: While the batch metrics help us iterate fast, a more representative metric is where for a given user, we perform a KNN search on the entire post corpora. We then average this metric over the entire set of users. While this metric is more computationally expensive, it is more indicative of online performance.

\section{Ablation studies}\label{sec:experiments}
We performed offline ablation studies to confirm the efficiency of every step. Our baseline model \cite{10.1145/2988450.2988454} uses deep \& wide architecture used with id features, where every item is assigned unique identifier and its embedding is trained as part of the model.
We trained a transformer architecture described in \S\,\ref{sec:model} with two separate sequences for pages and group posts, and static user features. It achieves +49\% absolute improvement in metrics.
Transformers bring substantial improvements into model quality. We were able to replace prior id based models in production. Additionally because the model is trained with pre-trained content embeddings and access to a large vocabulary, it can be applied to any new content which users enjoys over time.

We then added a CLS token - as a way to incorporate users with no prior engagement and we observed a lift in our offline metrics. We then switched over to causal mask. While this didn't give us any offline improvement, it enabled us to work on the future improvements to model a users short-term and long-term interests.

We targeted user's long-term interests by making the user embedding similar to multiple item embeddings from user's engagement future. By doing this, we target user's multiple interests and also make the model work consistently over a period of time.

With all techniques together we are able to achieve over 49\% absolute improvement over the wide and deep model baseline. Biggest wins comes from Two Tower Transformer and modeling user's long-term interests. We observe that {\systemname} system of a large vocabulary transformer based approach is able to distill dynamic nature of user history, retrieve engaging recommendations and out perform prior state of the art approaches. You can read our offline ablation study in the table \ref{tab:ablation_study_model}.

\begin{table}[tb]
\begin{tabular}{c|c}
    \toprule
    Technique & Batch Hits@1 \\
    \midrule
    Wide \& dense baseline \cite{10.1145/2988450.2988454} & 0.15 \\
    Two Tower Transformer (TTT) & 0.44 (+29\%)\\
    Two Tower Transformer (TTT) + CLS & 0.46 (+31\%)\\
    Above with Causal Mask & 0.46 (+31\%) \\
    Above + Long-term & 0.60 (+45\%) \\
    Above + Short-Term & 0.62 (+47\%)  \\
    Above + Relative Time Feature & 0.64 (+49\%) \\
    \bottomrule
\end{tabular}
\vspace{2mm}
\caption{Ablation study of different model configuration given same training and evaluation data.}
\label{tab:ablation_study_model}
%\vspace{-2.0em}
\end{table}

\begin{figure}[tb]
  \centering
  \includegraphics[width=\linewidth]{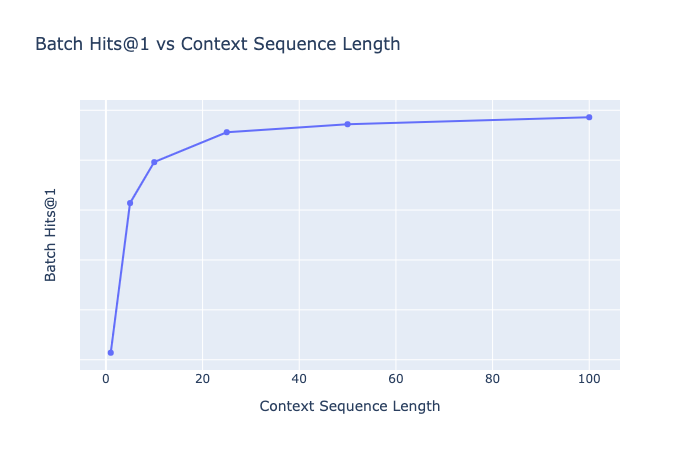}
  \vspace{-2em}
  \caption{Changes in Hits@1 given the sequence length. Experimented with Two Tower Transformer variant from Table \ref{tab:ablation_study_model}.}
  \label{fig:batch_length}
  \vspace{-1.0em}
\end{figure}

\begin{figure}[tb]
  \centering
  \includegraphics[width=\linewidth]{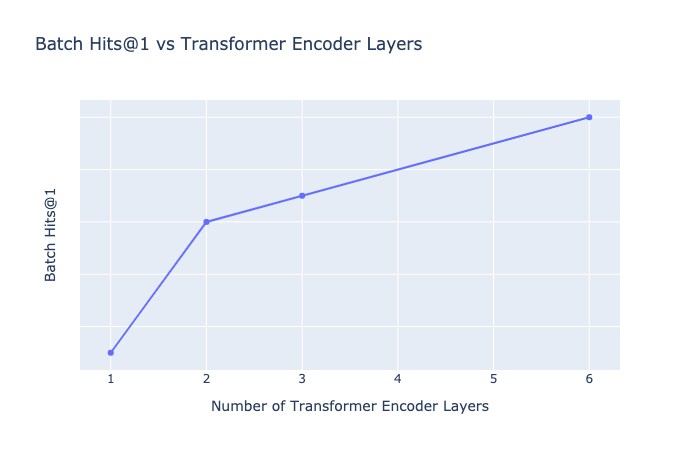}
  \vspace{-2em}
  \caption{Changes in Hits@1 with the number of transformer encoder layers.}
  \label{fig:batch_encoder_layers}
  \vspace{-1.0em}
\end{figure}

\subsection{Varying Sequence Length and Encoder Layers}\label{sec:sequence_length_and_layers}
To understand the model performance we analysed the sequence length required to improve model performance. We note that the longer the sequence length the higher to computational costs of the model. As shown in Fig.~\ref{fig:batch_length} beyond a sequence length of around 50-100 we begin to see diminishing returns. We have use similar sequence length for groups and page posts. 
We did a data analysis to understand, which sequence length would be the effective considering the user history. We have used a content understanding model for comparing a post in the history to the target post clicked in the future.

We tuned number of layers in the encoder and choose two layers (Fig. \ref{fig:batch_encoder_layers}), because computational load is increasing more rapidly for number of layers in comparison to improvement in Hits@1 we are getting. We tuned both batch length and number of encoder layers based on \textit{Two Tower Transformer} variant from Table \ref{tab:ablation_study_model}.

\subsection{Using relative time of engagement as a feature}
We observed that adding the relative time of engagement is a useful feature to capture user behavior. Some users prefer to engage with a lot of posts in a short period of time whereas some users engage with posts after more than a week. By incorporating this feature into the model, we have learnt the following lessons from our offline study - (i) the model gives less preferences to older engagement regardless of the position in the history. (ii) we observe that the model pays more attention to posts, where user signaled more long term interest such that visiting the group and consuming posts across multiple user sessions. We've summarized our offline ablation analysis in the table \ref{tab:engagement_time}.

\begin{table}[tb]
\begin{tabular}{c|c}
    \toprule
    Technique & Batch Hits@1 \\
    \midrule
    {\systemname} Model & baseline \\
    {\systemname} Model without engagement feature &  -1.6\% \\
    \bottomrule
\end{tabular}
\vspace{2mm}
\caption{Ablation study of the relative time of engagement feature.}
\label{tab:engagement_time}
\vspace{-2.0em}
\end{table}

\subsection{Varying Number of Negatives}

\begin{table}[tb]
\begin{tabular}{c|c}
    \toprule
    Technique & Batch Hits@1 \\
    \midrule
    {\systemname} Model & baseline \\
    {\systemname} model with 500 sampled negatives  & -5.3\% \\
    {\systemname} model with 1000 sampled negatives  & -3.1\% \\
    \bottomrule
\end{tabular}
\vspace{2mm}
\caption{Ablation study of the number of negatives for both the short-term and long-term losses.}
\label{tab:num_neg}
\vspace{-2.0em}
\end{table}

Fetching in-batch negatives in a standard TwoTower model is a standard task - For the user i, item at index i is the positive and all the other items in the batch are negatives. However, we've 2 tasks and both have more than 1 positives and hence we describe our process of fetching negatives. For modeling short-term interests, we label it as a multi-label multi-class problem. For a given user, we could have upto N positives (where N is the maximum sequence length in the Transformer). We use all the posts from other users as the negative pool. For modeling long-term interests, we've multiple labels per user as positive. We again model this as a multi-label multi-class problem and use all the label posts from other users as negatives. 

In this section, we explore an option of uniformly sampling from the pool and use only a fraction of them as negatives. From the negative pool, we uniformly sample (500, 1000) posts and use them as negatives. However, we found that using the entire pool always work best for us. We've summarized our results in Table \ref{tab:num_neg}.

\subsection{Dealing with Cold-start and Marginal Users}

As {\systemname} is a sequential model, it doesn't perform well for cold-start (users with no past engagement) and marginal (users with just a handful of past engagement) users. In order to deal with this, we took a couple of approaches - backfilling user engagement with popular posts and using a personalized model to backfill engagement from other similar users. In the first approach, we got the most popular posts and used this as a user history. In order to improve relevance, we only selected posts having the same attributes (like location, language) as the user. This helps the cold-start users. In the second approach, we built a user-user similarity model based on engagement and backfilled the history of marginal users from other similar users. This helps in improving the metrics for marginal users and we've summarized our results in Table \ref{tab:cold-start}. Note that this evaluation dataset only contains marginal and cold-start users and is different from the rest of the evaluation dataset.

\begin{table}[tb]
\begin{tabular}{c|c}
    \toprule
    Technique & Batch Hits@1 \\
    \midrule
    {\systemname} Model & baseline \\
    Sequence with popular posts  & +3.2\% \\
    Personalized Model  & +6.1\% \\
    \bottomrule
\end{tabular}
\vspace{2mm}
\caption{Offline metrics only for cold-start and marginal users.}
\label{tab:cold-start}
\vspace{-2.0em}
\end{table}

\section{System architecture}\label{sec:architecture}
\begin{figure}[tb]
  \centering
  \includegraphics[width=8cm]{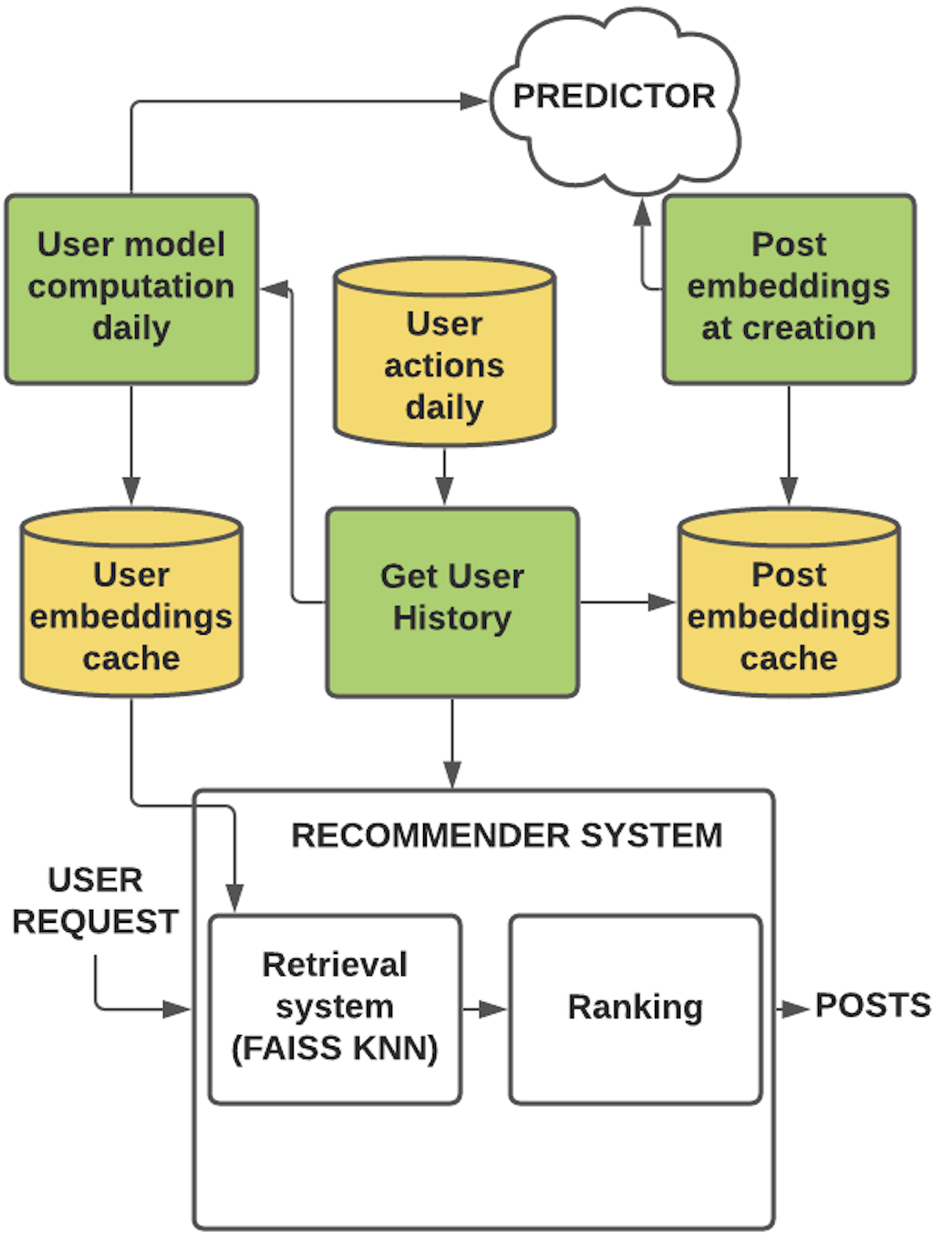}
  %\vspace{-2em}
  \caption{System architecture of {\systemname}.}
  \label{fig:architecture_pic}
  %\vspace{-1.0em}
\end{figure}

We illustrate the system architecture in Fig.~\ref{fig:architecture_pic}.
{\systemname} is deployed in production and is designed to operate on products created in real-time. Model inferences are computed at cloud of machines called Predictor~\cite{predictor_paper}. Predictor provides functionality to deploy the model, and API to call the model with a set of input features. Predictor is a cloud service and can scale accordingly.

\subsection{Serving Post Embeddings} 
Upon the creation and update of a post, an asynchronous call will be made to Predictor to generate its post embedding, and the embedding vector will be updated to recommendations index in real-time to prepare the post for retrieval and ranking. In other words, post embeddings are already pre-computed and indexed when search queries are issued, which makes the computation of user to post similarity across many post candidates tractable. For purposes of user history inference post side embeddings are also stored to distributed file system. We access post embeddings from distributed file system when we rerun the user embeddings refresh inference job.

\subsection{Serving User Embeddings}
We designed re-computation of user embeddings daily for the users who have fresh engagements with a content posts in Facebook ecosystem. We collect engaged posts ids, and join it with embeddings from the embedding cache in distributed file system and provide most recent user history to User Tower model, which recompute fresh user embeddings. We filter out integrity violating posts from the post ids list according to defined safety procedures \cite{GroupsIntegrity}.
After re-computation we refresh the user embeddings in key-value distributed memory lookup store. 

\subsection{Recommendation candidates retrieval}
At Facebook, our systems need to handle a large QPS to serve our large user base. Queries to recommendations system are requested in real time, and given user id, we retrieve its pre-computed embedding from key-value store in milliseconds. Given user embeddings within {\systemname} resulted posts are retrieved from recommendations index in less than several hundred milliseconds.
Our recommendation retrieval engine uses \cite{faiss_paper} to compress the embedding space to speed up the k-nearest neighbor computations.
Results are then passed back to several stages of ranking.

\section{Deployment lessons}\label{sec:deployment}
\begin{table}[tb]
\begin{tabular}{c|c}
    \toprule
    Technique & Online results  \\
    \midrule
    Two Tower Transformer (TTT) & +0.18\% \\
    Two Tower Transformer + long-range  & +0.5\% \\
    Two Tower Transformer + short \& long-range  & +0.6\% \\
    \bottomrule
\end{tabular}
\vspace{2mm}
\caption{Online A/B testing results based on model variants in Table \ref{tab:ablation_study_model} for relative improvement in number of groups users who are meeting new people, engaging with the community, sharing knowledge and getting support.}
\label{tab:online_results}
\vspace{-2.0em}
\end{table}

We deployed {\systemname} on Facebook Groups recommendation system powering billions of queries per day. In our first iteration, we deployed the vanilla \textit{Two Tower Transformer} (Table~\ref{tab:ablation_study_model}), which achieves 0.44 Hits@1 in the batch. We observed relative increase of 0.18\%  in number of users, who used groups to meeting new people, engaging with the community, sharing knowledge and getting support. In the next iteration, we improved the model substantially by learning user's long-term interests with a relative improvement of 0.5\%. We further improved upon this to 0.6\% by learning both the short-term and long-term interests. We learnt several important deployment lessons while during development of {\systemname}.

\subsection{User History Freshness}
In our online experiments, we have enabled daily refresh of user embeddings by updating their context. With this setup, it naturally begs the question - Is it necessary to update the user context daily? What is the loss in metric by not doing this? We try to answer these questions analytically through both online and offline experiments. In our offline experiments, we compare Recall@20 when the context is fresh with the context being stale. 

\begin{figure}[tb]
  \centering
  %\vspace{-0.1em}
  \includegraphics[width=\linewidth]{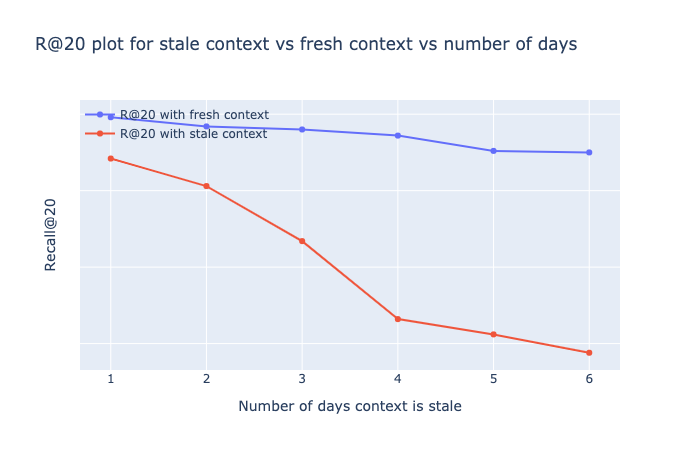}
 % \vspace{-2em}
  \caption{We find that Recall@20 drops by 3\% if the history is not updated for a day, 5\% if they are not updated for 2 days and up to 20\% if they're not updated for 6 days.}
  \label{fig:engagement_offline_drop}
  % \vspace{-1.0em}
\end{figure}

We observed that if we do not refresh the user embeddings daily then both offline (Fig.~\ref{fig:engagement_offline_drop}) and online metrics regress and go down day by day. This is intuitive because user's interests change over time and also user can not click $like$ on the same post twice, so showing the fresh content based on what user interacted with recently is important to continue to entertain the user.

\subsection{Model performance over time}

During the training data collection process, it is important that the (user context, target post) pair are not restricted to a small time period (like a day or a week). The reason for this is that posts might be biased towards a set of particular topics and we don't want the model to be biased towards these topics. In order to test our hypothesis, we trained our model with the target post coming from a short time period and ran offline KNN evaluation over the next 7 days. We observed that the KNN Hits@20 drops by around 30\% over a period of 7 days.

We solve this problem by modeling user's long term interests. The main intuition behind modeling user's long-term interests is that the user embedding is learned to match with their interests over the next few days. After introducing the long term interests, we observed that the KNN Hits@20 drops by only 2.3\% at the 7 day period, as opposed to 30\% before.

\subsection{Targeting and Precision}
It is important for recommendation system to stays highly engaged. We measure system efficiency by click-through rate (CTR), which means that ratio of \textit{number of user actions to number of user impressions of recommended posts} need to stay constant or improve over time of system development. To make sure {\systemname}'s User to post embeddings improved or do not regress click-through rate we implemented filtering of returned results by a threshold. This helps to move from -6.2\% in CTR  regression to neutral CTR movement keeping the system more efficient by recommending more related content to users based on their interests.

\section{Conclusion}\label{sec:conclusion}
In this paper we presented Sequence based User to Post recommendation system called {\systemname}. We proposed to extend Transformer Based Sequence model to support pre-trained item embeddings, a large vocabulary and use two tower architecture instead of a classifier. We use a causal multi-head attention to learn both user's short-term and long-term interests. With all of the techniques explored we are able to increase model performance by over 49\% absolute in comparison to baseline. We observed that as a result of {\systemname} deployment, 0.6\% more users are meeting new people, engaging with the community, sharing knowledge and getting support.

In the future iterations of the {\systemname} we plan to explore near-real time model inference. We are also looking to explore Graph Neural Networks and bringing in other entities like groups, pages, search queries and hashtags.

\section{Acknowledgements}\label{sec:acknowledgements}
The authors would like to thank Shaoliang Nie, Ignacio Arranz, Liang Tan, Hamed Firooz, Nikhil Garg, Qing Xu, Jun Mei, Sheng Song, Zhen Li and others who contributed, supported and collaborated with us.

\balance
\bibliographystyle{ACM-Reference-Format}
\bibliography{bibliography}


\begin{thebibliography}{29}

%%% the Web address.

\ifx \showCODEN    \undefined \def \showCODEN     #1{\unskip}     \fi
\ifx \showDOI      \undefined \def \showDOI       #1{#1}\fi
\ifx \showISBNx    \undefined \def \showISBNx     #1{\unskip}     \fi
\ifx \showISBNxiii \undefined \def \showISBNxiii  #1{\unskip}     \fi
\ifx \showISSN     \undefined \def \showISSN      #1{\unskip}     \fi
\ifx \showLCCN     \undefined \def \showLCCN      #1{\unskip}     \fi
\ifx \shownote     \undefined \def \shownote      #1{#1}          \fi
\ifx \showarticletitle \undefined \def \showarticletitle #1{#1}   \fi
\ifx \showURL      \undefined \def \showURL       {\relax}        \fi
% invisible to TeX
\providecommand\bibfield[2]{#2}
\providecommand\bibinfo[2]{#2}
\providecommand\natexlab[1]{#1}
\providecommand\showeprint[2][]{arXiv:#2}

\bibitem[\protect\citeauthoryear{Alison}{Alison}{2021}]%
        {GroupsIntegrity}
\bibfield{author}{\bibinfo{person}{Tom Alison}.}
  \bibinfo{year}{2021}\natexlab{}.
\newblock \bibinfo{title}{Changes to Keep Facebook Groups Safe}.
\newblock
\newblock
\urldef\tempurl%
\url{https://about.fb.com/news/2021/03/changes-to-keep-facebook-groups-safe/}
\showURL{%
\tempurl}


\bibitem[\protect\citeauthoryear{Borisyuk, Malreddy, Mei, Liu, Liu, Maheshwari,
  Bell, and Rangadurai}{Borisyuk et~al\mbox{.}}{2021}]%
        {visrel_paper}
\bibfield{author}{\bibinfo{person}{Fedor Borisyuk}, \bibinfo{person}{Siddarth
  Malreddy}, \bibinfo{person}{Jun Mei}, \bibinfo{person}{Yiqun Liu},
  \bibinfo{person}{Xiaoyi Liu}, \bibinfo{person}{Piyush Maheshwari},
  \bibinfo{person}{Anthony Bell}, {and} \bibinfo{person}{Kaushik Rangadurai}.}
  \bibinfo{year}{2021}\natexlab{}.
\newblock \showarticletitle{VisRel: Media Search at Scale}. In
  \bibinfo{booktitle}{\emph{KDD}}.
\newblock


\bibitem[\protect\citeauthoryear{Cheng, Koc, Harmsen, Shaked, Chandra, Aradhye,
  Anderson, Corrado, Chai, Ispir, Anil, Haque, Hong, Jain, Liu, and Shah}{Cheng
  et~al\mbox{.}}{2016}]%
        {10.1145/2988450.2988454}
\bibfield{author}{\bibinfo{person}{Heng-Tze Cheng}, \bibinfo{person}{Levent
  Koc}, \bibinfo{person}{Jeremiah Harmsen}, \bibinfo{person}{Tal Shaked},
  \bibinfo{person}{Tushar Chandra}, \bibinfo{person}{Hrishi Aradhye},
  \bibinfo{person}{Glen Anderson}, \bibinfo{person}{Greg Corrado},
  \bibinfo{person}{Wei Chai}, \bibinfo{person}{Mustafa Ispir},
  \bibinfo{person}{Rohan Anil}, \bibinfo{person}{Zakaria Haque},
  \bibinfo{person}{Lichan Hong}, \bibinfo{person}{Vihan Jain},
  \bibinfo{person}{Xiaobing Liu}, {and} \bibinfo{person}{Hemal Shah}.}
  \bibinfo{year}{2016}\natexlab{}.
\newblock \showarticletitle{Wide \& Deep Learning for Recommender Systems}. In
  \bibinfo{booktitle}{\emph{RecSys}}.
\newblock


\bibitem[\protect\citeauthoryear{Conneau, wal, Goyal, Chaudhary, Wenzek,
  Guzm{\'{a}}n, Grave, Ott, Zettlemoyer, and Stoyanov}{Conneau
  et~al\mbox{.}}{2020}]%
        {DBLP:conf/acl/ConneauKGCWGGOZ20}
\bibfield{author}{\bibinfo{person}{Alexis Conneau},
  \bibinfo{person}{Kartikay~Khandel wal}, \bibinfo{person}{Naman Goyal},
  \bibinfo{person}{Vishrav Chaudhary}, \bibinfo{person}{Guillaume Wenzek},
  \bibinfo{person}{Francisco Guzm{\'{a}}n}, \bibinfo{person}{Edouard Grave},
  \bibinfo{person}{Myle Ott}, \bibinfo{person}{Luke Zettlemoyer}, {and}
  \bibinfo{person}{Veselin Stoyanov}.} \bibinfo{year}{2020}\natexlab{}.
\newblock \showarticletitle{Unsupervised Cross-lingual Representation Learning
  at Scale}. In \bibinfo{booktitle}{\emph{ACL}}.
\newblock


\bibitem[\protect\citeauthoryear{Covington, Adams, and Sargin}{Covington
  et~al\mbox{.}}{2016}]%
        {10.1145/2959100.2959190}
\bibfield{author}{\bibinfo{person}{Paul Covington}, \bibinfo{person}{Jay
  Adams}, {and} \bibinfo{person}{Emre Sargin}.}
  \bibinfo{year}{2016}\natexlab{}.
\newblock \showarticletitle{Deep Neural Networks for YouTube Recommendations}.
  In \bibinfo{booktitle}{\emph{RecSys}}.
\newblock


\bibitem[\protect\citeauthoryear{Deng, Guo, Xue, and Zafeiriou}{Deng
  et~al\mbox{.}}{2019}]%
        {deng2019arcface}
\bibfield{author}{\bibinfo{person}{Jiankang Deng}, \bibinfo{person}{Jia Guo},
  \bibinfo{person}{Niannan Xue}, {and} \bibinfo{person}{Stefanos Zafeiriou}.}
  \bibinfo{year}{2019}\natexlab{}.
\newblock \bibinfo{title}{ArcFace: Additive Angular Margin Loss for Deep Face
  Recognition}.
\newblock
\newblock
\showeprint[arxiv]{cs.CV/1801.07698}


\bibitem[\protect\citeauthoryear{{Hazelwood}, {Bird}, {Brooks}, {Chintala},
  {Diril}, {Dzhulgakov}, {Fawzy}, {Jia}, {Jia}, {Kalro}, {Law}, {Lee}, {Lu},
  {Noordhuis}, {Smelyanskiy}, {Xiong}, and {Wang}}{{Hazelwood}
  et~al\mbox{.}}{2018}]%
        {predictor_paper}
\bibfield{author}{\bibinfo{person}{K. {Hazelwood}}, \bibinfo{person}{S.
  {Bird}}, \bibinfo{person}{D. {Brooks}}, \bibinfo{person}{S. {Chintala}},
  \bibinfo{person}{U. {Diril}}, \bibinfo{person}{D. {Dzhulgakov}},
  \bibinfo{person}{M. {Fawzy}}, \bibinfo{person}{B. {Jia}}, \bibinfo{person}{Y.
  {Jia}}, \bibinfo{person}{A. {Kalro}}, \bibinfo{person}{J. {Law}},
  \bibinfo{person}{K. {Lee}}, \bibinfo{person}{J. {Lu}}, \bibinfo{person}{P.
  {Noordhuis}}, \bibinfo{person}{M. {Smelyanskiy}}, \bibinfo{person}{L.
  {Xiong}}, {and} \bibinfo{person}{X. {Wang}}.}
  \bibinfo{year}{2018}\natexlab{}.
\newblock \showarticletitle{Applied Machine Learning at Facebook: A Datacenter
  Infrastructure Perspective}. In \bibinfo{booktitle}{\emph{HPCA}}.
\newblock


\bibitem[\protect\citeauthoryear{Huang, He, Gao, Deng, Acero, and Heck}{Huang
  et~al\mbox{.}}{2013}]%
        {huang2013learning}
\bibfield{author}{\bibinfo{person}{Po-Sen Huang}, \bibinfo{person}{Xiaodong
  He}, \bibinfo{person}{Jianfeng Gao}, \bibinfo{person}{Li Deng},
  \bibinfo{person}{Alex Acero}, {and} \bibinfo{person}{Larry Heck}.}
  \bibinfo{year}{2013}\natexlab{}.
\newblock \showarticletitle{Learning deep structured semantic models for web
  search using clickthrough data}. In \bibinfo{booktitle}{\emph{CIKM}}.
\newblock


\bibitem[\protect\citeauthoryear{Jean, Cho, Memisevic, and Bengio}{Jean
  et~al\mbox{.}}{2014}]%
        {DBLP:journals/corr/JeanCMB14}
\bibfield{author}{\bibinfo{person}{S{\'{e}}bastien Jean},
  \bibinfo{person}{Kyunghyun Cho}, \bibinfo{person}{Roland Memisevic}, {and}
  \bibinfo{person}{Yoshua Bengio}.} \bibinfo{year}{2014}\natexlab{}.
\newblock \showarticletitle{On Using Very Large Target Vocabulary for Neural
  Machine Translation}.
\newblock \bibinfo{journal}{\emph{CoRR}} (\bibinfo{year}{2014}).
\newblock


\bibitem[\protect\citeauthoryear{{Johnson}, {Douze}, and {Jégou}}{{Johnson}
  et~al\mbox{.}}{2019}]%
        {faiss_paper}
\bibfield{author}{\bibinfo{person}{J. {Johnson}}, \bibinfo{person}{M. {Douze}},
  {and} \bibinfo{person}{H. {Jégou}}.} \bibinfo{year}{2019}\natexlab{}.
\newblock \showarticletitle{Billion-scale similarity search with GPUs}.
\newblock \bibinfo{journal}{\emph{IEEE Transactions on Big Data}}
  (\bibinfo{year}{2019}).
\newblock


\bibitem[\protect\citeauthoryear{Kang and McAuley}{Kang and McAuley}{2018}]%
        {Kang2018SelfAttentiveSR}
\bibfield{author}{\bibinfo{person}{Wang-Cheng Kang} {and}
  \bibinfo{person}{Julian McAuley}.} \bibinfo{year}{2018}\natexlab{}.
\newblock \showarticletitle{Self-Attentive Sequential Recommendation}.
\newblock \bibinfo{journal}{\emph{2018 IEEE International Conference on Data
  Mining (ICDM)}} (\bibinfo{year}{2018}), \bibinfo{pages}{197--206}.
\newblock


\bibitem[\protect\citeauthoryear{Kipf and Welling}{Kipf and Welling}{2017}]%
        {DBLP:conf/iclr/KipfW17}
\bibfield{author}{\bibinfo{person}{Thomas~N. Kipf} {and} \bibinfo{person}{Max
  Welling}.} \bibinfo{year}{2017}\natexlab{}.
\newblock \showarticletitle{Semi-Supervised Classification with Graph
  Convolutional Networks}. In \bibinfo{booktitle}{\emph{ICLR}}.
\newblock


\bibitem[\protect\citeauthoryear{Kislyuk, Liu, Liu, Tzeng, and Jing}{Kislyuk
  et~al\mbox{.}}{2015}]%
        {kislyuk2015human}
\bibfield{author}{\bibinfo{person}{Dmitry Kislyuk}, \bibinfo{person}{Yuchen
  Liu}, \bibinfo{person}{David Liu}, \bibinfo{person}{Eric Tzeng}, {and}
  \bibinfo{person}{Yushi Jing}.} \bibinfo{year}{2015}\natexlab{}.
\newblock \bibinfo{title}{Human Curation and Convnets: Powering Item-to-Item
  Recommendations on Pinterest}.
\newblock
\newblock


\bibitem[\protect\citeauthoryear{Lerer, Wu, Shen, Lacroix, Wehrstedt, Bose, and
  Peysakhovich}{Lerer et~al\mbox{.}}{2019}]%
        {MLSYS2019_e2c420d9}
\bibfield{author}{\bibinfo{person}{Adam Lerer}, \bibinfo{person}{Ledell Wu},
  \bibinfo{person}{Jiajun Shen}, \bibinfo{person}{Timothee Lacroix},
  \bibinfo{person}{Luca Wehrstedt}, \bibinfo{person}{Abhijit Bose}, {and}
  \bibinfo{person}{Alex Peysakhovich}.} \bibinfo{year}{2019}\natexlab{}.
\newblock \showarticletitle{Pytorch-BigGraph: A Large Scale Graph Embedding
  System}. In \bibinfo{booktitle}{\emph{MLSys}}.
\newblock


\bibitem[\protect\citeauthoryear{Linden, Smith, and York}{Linden
  et~al\mbox{.}}{2003}]%
        {10.1109/MIC.2003.1167344}
\bibfield{author}{\bibinfo{person}{Greg Linden}, \bibinfo{person}{Brent Smith},
  {and} \bibinfo{person}{Jeremy York}.} \bibinfo{year}{2003}\natexlab{}.
\newblock \showarticletitle{Amazon.com Recommendations: Item-to-Item
  Collaborative Filtering}.
\newblock \bibinfo{journal}{\emph{IEEE Internet Computing}}
  (\bibinfo{year}{2003}).
\newblock


\bibitem[\protect\citeauthoryear{Liu, Rangadurai, He, Malreddy, Gui, Liu, and
  Borisyuk}{Liu et~al\mbox{.}}{2021}]%
        {que2search_paper}
\bibfield{author}{\bibinfo{person}{Yiqun Liu}, \bibinfo{person}{Kaushik
  Rangadurai}, \bibinfo{person}{Yunzhong He}, \bibinfo{person}{Siddarth
  Malreddy}, \bibinfo{person}{Xunlong Gui}, \bibinfo{person}{Xiaoyi Liu}, {and}
  \bibinfo{person}{Fedor Borisyuk}.} \bibinfo{year}{2021}\natexlab{}.
\newblock \showarticletitle{Que2Search: Fast and Accurate Query and Document
  Understanding for Search at Facebook}. In \bibinfo{booktitle}{\emph{KDD}}.
\newblock


\bibitem[\protect\citeauthoryear{Niu, Li, Li, Xiao, Sun, Deng, and Chen}{Niu
  et~al\mbox{.}}{2020}]%
        {alibaba_two_tower}
\bibfield{author}{\bibinfo{person}{Xichuan Niu}, \bibinfo{person}{Bofang Li},
  \bibinfo{person}{Chenliang Li}, \bibinfo{person}{Rong Xiao},
  \bibinfo{person}{Haochuan Sun}, \bibinfo{person}{Hongbo Deng}, {and}
  \bibinfo{person}{Zhenzhong Chen}.} \bibinfo{year}{2020}\natexlab{}.
\newblock \showarticletitle{A Dual Heterogeneous Graph Attention Network to
  Improve Long-Tail Performance for Shop Search in E-Commerce}. In
  \bibinfo{booktitle}{\emph{KDD}}.
\newblock


\bibitem[\protect\citeauthoryear{Room}{Room}{2019}]%
        {fbgroup_stats}
\bibfield{author}{\bibinfo{person}{Facebook~News Room}.}
  \bibinfo{year}{2019}\natexlab{}.
\newblock \bibinfo{title}{The next step in Facebook’s AI hardware
  infrastructure}.
\newblock
\newblock
\newblock
\shownote{\url{https://about.fb.com/news/2019/04/f8-2019-day-1/}.}


\bibitem[\protect\citeauthoryear{Salakhutdinov, Mnih, and Hinton}{Salakhutdinov
  et~al\mbox{.}}{2007}]%
        {10.1145/1273496.1273596}
\bibfield{author}{\bibinfo{person}{Ruslan Salakhutdinov},
  \bibinfo{person}{Andriy Mnih}, {and} \bibinfo{person}{Geoffrey Hinton}.}
  \bibinfo{year}{2007}\natexlab{}.
\newblock \showarticletitle{Restricted Boltzmann Machines for Collaborative
  Filtering}. In \bibinfo{booktitle}{\emph{ICML}}.
\newblock


\bibitem[\protect\citeauthoryear{Sarwar, Karypis, Konstan, and Riedl}{Sarwar
  et~al\mbox{.}}{2001}]%
        {10.1145/371920.372071}
\bibfield{author}{\bibinfo{person}{Badrul Sarwar}, \bibinfo{person}{George
  Karypis}, \bibinfo{person}{Joseph Konstan}, {and} \bibinfo{person}{John
  Riedl}.} \bibinfo{year}{2001}\natexlab{}.
\newblock \showarticletitle{Item-Based Collaborative Filtering Recommendation
  Algorithms}. In \bibinfo{booktitle}{\emph{WWW}}.
\newblock


\bibitem[\protect\citeauthoryear{Saveski and Mantrach}{Saveski and
  Mantrach}{2014}]%
        {10.1145/2645710.2645751}
\bibfield{author}{\bibinfo{person}{Martin Saveski} {and} \bibinfo{person}{Amin
  Mantrach}.} \bibinfo{year}{2014}\natexlab{}.
\newblock \showarticletitle{Item Cold-Start Recommendations: Learning Local
  Collective Embeddings}. In \bibinfo{booktitle}{\emph{RecSys}}.
\newblock


\bibitem[\protect\citeauthoryear{Sun, Liu, Wu, Pei, Lin, Ou, and Jiang}{Sun
  et~al\mbox{.}}{2019}]%
        {10.1145/3357384.3357895}
\bibfield{author}{\bibinfo{person}{Fei Sun}, \bibinfo{person}{Jun Liu},
  \bibinfo{person}{Jian Wu}, \bibinfo{person}{Changhua Pei},
  \bibinfo{person}{Xiao Lin}, \bibinfo{person}{Wenwu Ou}, {and}
  \bibinfo{person}{Peng Jiang}.} \bibinfo{year}{2019}\natexlab{}.
\newblock \showarticletitle{BERT4Rec: Sequential Recommendation with
  Bidirectional Encoder Representations from Transformer}. In
  \bibinfo{booktitle}{\emph{CIKM}}.
\newblock


\bibitem[\protect\citeauthoryear{Tang, Belletti, Jain, Chen, Beutel, Xu, and
  H.~Chi}{Tang et~al\mbox{.}}{2019}]%
        {10.1145/3308558.3313650}
\bibfield{author}{\bibinfo{person}{Jiaxi Tang}, \bibinfo{person}{Francois
  Belletti}, \bibinfo{person}{Sagar Jain}, \bibinfo{person}{Minmin Chen},
  \bibinfo{person}{Alex Beutel}, \bibinfo{person}{Can Xu}, {and}
  \bibinfo{person}{Ed H.~Chi}.} \bibinfo{year}{2019}\natexlab{}.
\newblock \showarticletitle{Towards Neural Mixture Recommender for Long Range
  Dependent User Sequences}. In \bibinfo{booktitle}{\emph{WWW}}.
\newblock


\bibitem[\protect\citeauthoryear{Wang, Tran, and Feiszli}{Wang
  et~al\mbox{.}}{2019}]%
        {video_rep_paper3}
\bibfield{author}{\bibinfo{person}{Weiyao Wang}, \bibinfo{person}{Du Tran},
  {and} \bibinfo{person}{Matt Feiszli}.} \bibinfo{year}{2019}\natexlab{}.
\newblock \showarticletitle{What Makes Training Multi-Modal Networks Hard?}
\newblock \bibinfo{journal}{\emph{CVPR}} (\bibinfo{year}{2019}).
\newblock


\bibitem[\protect\citeauthoryear{Yang, Yi, Zhiyuan~Cheng, Hong, Li,
  Xiaoming~Wang, Xu, and Chi}{Yang et~al\mbox{.}}{2020}]%
        {10.1145/3366424.3386195}
\bibfield{author}{\bibinfo{person}{Ji Yang}, \bibinfo{person}{Xinyang Yi},
  \bibinfo{person}{Derek Zhiyuan~Cheng}, \bibinfo{person}{Lichan Hong},
  \bibinfo{person}{Yang Li}, \bibinfo{person}{Simon Xiaoming~Wang},
  \bibinfo{person}{Taibai Xu}, {and} \bibinfo{person}{Ed~H. Chi}.}
  \bibinfo{year}{2020}\natexlab{}.
\newblock \showarticletitle{Mixed Negative Sampling for Learning Two-Tower
  Neural Networks in Recommendations}. In \bibinfo{booktitle}{\emph{WWW}}.
\newblock


\bibitem[\protect\citeauthoryear{Yi, Yang, Hong, Cheng, Heldt, Kumthekar, Zhao,
  Wei, and Chi}{Yi et~al\mbox{.}}{2019}]%
        {10.1145/3298689.3346996}
\bibfield{author}{\bibinfo{person}{Xinyang Yi}, \bibinfo{person}{Ji Yang},
  \bibinfo{person}{Lichan Hong}, \bibinfo{person}{Derek~Zhiyuan Cheng},
  \bibinfo{person}{Lukasz Heldt}, \bibinfo{person}{Aditee Kumthekar},
  \bibinfo{person}{Zhe Zhao}, \bibinfo{person}{Li Wei}, {and}
  \bibinfo{person}{Ed Chi}.} \bibinfo{year}{2019}\natexlab{}.
\newblock \showarticletitle{Sampling-Bias-Corrected Neural Modeling for Large
  Corpus Item Recommendations}. In \bibinfo{booktitle}{\emph{RecSys}}.
\newblock


\bibitem[\protect\citeauthoryear{Ying, He, Chen, Eksombatchai, Hamilton, and
  Leskovec}{Ying et~al\mbox{.}}{2018}]%
        {10.1145/3219819.3219890}
\bibfield{author}{\bibinfo{person}{Rex Ying}, \bibinfo{person}{Ruining He},
  \bibinfo{person}{Kaifeng Chen}, \bibinfo{person}{Pong Eksombatchai},
  \bibinfo{person}{William~L. Hamilton}, {and} \bibinfo{person}{Jure
  Leskovec}.} \bibinfo{year}{2018}\natexlab{}.
\newblock \showarticletitle{Graph Convolutional Neural Networks for Web-Scale
  Recommender Systems}. In \bibinfo{booktitle}{\emph{KDD}}.
\newblock


\bibitem[\protect\citeauthoryear{Zaheer, Kottur, Ravanbakhsh, P{\'{o}}czos,
  Salakhutdinov, and Smola}{Zaheer et~al\mbox{.}}{2017}]%
        {DBLP:journals/corr/ZaheerKRPSS17}
\bibfield{author}{\bibinfo{person}{Manzil Zaheer}, \bibinfo{person}{Satwik
  Kottur}, \bibinfo{person}{Siamak Ravanbakhsh},
  \bibinfo{person}{Barnab{\'{a}}s P{\'{o}}czos}, \bibinfo{person}{Ruslan
  Salakhutdinov}, {and} \bibinfo{person}{Alexander~J. Smola}.}
  \bibinfo{year}{2017}\natexlab{}.
\newblock \showarticletitle{Deep Sets}.
\newblock \bibinfo{journal}{\emph{CoRR}} (\bibinfo{year}{2017}).
\newblock


\bibitem[\protect\citeauthoryear{Zhai}{Zhai}{2021}]%
        {pinterest_workshop}
\bibfield{author}{\bibinfo{person}{Andrew Zhai}.}
  \bibinfo{year}{2021}\natexlab{}.
\newblock \showarticletitle{Representation Learning for Recommender Systems}.
  In \bibinfo{booktitle}{\emph{KDD '21 OARS workshop}}.
\newblock


\end{thebibliography}

\end{document}